\documentclass[11pt,letterpaper]{article}

\usepackage[T1]{fontenc}

\usepackage{lmodern}

\usepackage{microtype}

\usepackage{graphicx}
\usepackage{amsmath}
\usepackage{amssymb}
\usepackage[round]{natbib}

\usepackage[
    letterpaper,
    textwidth=6.5in,
    textheight=9in,
    centering
]{geometry}

\usepackage{booktabs}
\usepackage{array}
\usepackage{tabularx}
\usepackage{multirow}

\usepackage[table,dvipsnames]{xcolor}

\usepackage{enumitem}
\usepackage{titlesec}
\usepackage{fancyhdr}
\usepackage{lastpage}
\usepackage{hyperref}
\usepackage{url}
\usepackage{subcaption}
\usepackage{tikz}
\usepackage{xspace}
\usetikzlibrary{positioning, arrows.meta, fit, calc, backgrounds}

\definecolor{AtlanticBlue}{HTML}{164A63}
\definecolor{OceanBlue}{HTML}{28799E}
\definecolor{MistBlue}{HTML}{EAF3F7}
\definecolor{SlateGrey}{HTML}{4B5963}
\definecolor{RuleGrey}{HTML}{CBD3D7}
\definecolor{WarmGrey}{HTML}{F4F4F2}

\definecolor{SSMFill}{HTML}{EDE4F7}
\definecolor{SSMEdge}{HTML}{7B2CBF}
\definecolor{AttFill}{HTML}{FDE8D7}
\definecolor{AttEdge}{HTML}{C2610A}
\definecolor{FFNFill}{HTML}{E4F0E2}
\definecolor{FFNEdge}{HTML}{4C7A3A}

\hypersetup{
  colorlinks=true,
  linkcolor=AtlanticBlue,
  citecolor=AtlanticBlue,
  urlcolor=OceanBlue,
  pdftitle={ZetaGPT: A Reference Positional-Encoding-Free State-Space-Attention Large Language Model},
  pdfauthor={Roisin Luo}
}

\renewcommand{\arraystretch}{1.14}

\setlist[itemize]{
  leftmargin=3pc,
  topsep=4pt plus 1pt minus 2pt,
  partopsep=1pt plus 0.5pt minus 0.5pt,
  itemsep=2pt plus 1pt minus 0.5pt,
  parsep=2pt plus 1pt minus 0.5pt
}

\setlist[enumerate]{
  leftmargin=3pc,
  topsep=4pt plus 1pt minus 2pt,
  partopsep=1pt plus 0.5pt minus 0.5pt,
  itemsep=2pt plus 1pt minus 0.5pt,
  parsep=2pt plus 1pt minus 0.5pt
}

\titleformat{\section}
  {\large\bfseries\color{AtlanticBlue}}
  {\thesection.}{0.45em}{}
  [\vspace{-2pt}\color{RuleGrey}\titlerule]
\titlespacing*{\section}{0pt}{8pt}{4pt}

\titleformat{\subsection}
  {\normalsize\bfseries\color{SlateGrey}}
  {\thesubsection.}{0.4em}{}
\titlespacing*{\subsection}{0pt}{6pt}{2pt}

\newcommand{\email}[1]{\href{mailto:#1}{\texttt{#1}}}

\renewcommand{\thanks}[1]{%
  \begingroup
    \renewcommand{\thefootnote}{\fnsymbol{footnote}}%
    \footnotemark[1]\footnotetext[1]{#1}%
  \endgroup
}

\newcolumntype{L}[1]{>{\raggedright\arraybackslash}p{#1}}
\newcolumntype{C}[1]{>{\centering\arraybackslash}p{#1}}
\newcolumntype{Y}{>{\raggedright\arraybackslash}X}

\RenewDocumentEnvironment{abstract}{+b}
{%
  \begin{center}
  \fcolorbox{RuleGrey}{MistBlue}{%
    \begin{minipage}{0.965\textwidth}
    \small
    \textbf{\color{AtlanticBlue}Abstract.}
    #1
    \end{minipage}%
  }%
  \end{center}
}
{}

\begin{document}

\begin{center}
  {\color{AtlanticBlue}\rule{\textwidth}{1.2pt}}\\[6pt]

  \begin{center}
    \begin{minipage}{1\linewidth}
    \centering
    {\LARGE\bfseries\color{AtlanticBlue}
    ZetaGPT: A Reference Implementation of Positional--Encoding--Free State--Space--Attention Language Models
    }
    \end{minipage}
  \end{center}

  \vspace{7pt}
  {\large R\'ois\'in Luo\thanks{Corresponding to \email{roisincrtai@gmail.com}}}

  \vspace{2pt}
  {\normalsize University of Galway, Ireland}

  \vspace{3pt}
  {\normalsize \url{https://github.com/roisincrtai/zetagpt}}

  \vspace{2pt}
  {\color{AtlanticBlue}\rule{\textwidth}{0.5pt}}
\end{center}

\begin{abstract}
Transformer-based language models rely on self-attention, whose computation is permutation-equivariant and therefore lacks an intrinsic mechanism for representing token order. Existing architectures address this limitation by explicitly incorporating positional information through learned positional embeddings or hand-crafted positional encodings, such as rotary positional encoding (RoPE), treating positional information as an architecturally acquired capability rather than an inherent property of the model. Motivated by the pursuit of positional-encoding-free architectures, this work explores a language model architecture that integrates causal state-space equations to implicitly encode positional information before attention computation. Specifically, each model block applies a causal state-space equation before self-attention, allowing recurrent state dynamics to encode sequential information into token representations. Consequently, subsequent attention layers operate on position-aware representations without requiring explicit positional encodings while retaining the expressive modeling capacity of self-attention. We present \textsc{ZetaGPT}, a compact hybrid language model designed for research, rapid prototyping, algorithm verification, and educational applications. In addition to the proposed architecture, \textsc{ZetaGPT} provides a fully open-source, end-to-end training pipeline encompassing dataset construction, tokenizer training, pretraining, supervised fine-tuning, reinforcement learning from human feedback (RLHF), and chain-of-thought (CoT) reasoning via pure reinforcement learning. To the best of our knowledge, \textsc{ZetaGPT} is the first open-source small language model without explicit positional encoding and establishes a compact, reproducible reference implementation for the development and empirical study of positional-encoding-free language models.
\end{abstract}

\vspace{3pt}

\textbf{\small Keywords:}
\small positional-encoding-free large language model; state-space model; state-space transformer; long context; pretraining; reinforcement learning from human feedback; direct preference optimization; large language model

\section{Introduction}

Large language models (LLMs) based on the Transformer architecture~\citep{vaswani2017attention} have demonstrated remarkable capabilities across a broad range of natural language processing tasks, including language understanding, reasoning, question answering, and code generation. Since GPT-2~\citep{radford2019gpt2}, the dominant paradigm has remained largely unchanged: a stack of Transformer blocks composed of multi-head self-attention and feed-forward networks, pretrained by next-token prediction on large-scale text corpora and subsequently adapted through supervised fine-tuning and alignment. Despite rapid advances in model scale and training methodology, the fundamental architecture of self-attention remains the backbone of modern LLMs.

A fundamental property of self-attention is permutation equivariance: applying any permutation $\pi$ to an input sequence results in the same permutation of the output representations. Formally, for an input sequence $\mathbf{x}=(x_1,x_2,\ldots,x_T)$ of length $T$,
\begin{align}
    \mathrm{Attn}\!\left(\pi(\mathbf{x})\right)
    =
    \pi\!\left(\mathrm{Attn}(\mathbf{x})\right),
\end{align}
where $\mathrm{Attn}(\cdot)$ denotes the self-attention operation and $\pi$ is an arbitrary permutation of the input sequence. Consequently, self-attention preserves the permutation structure of its inputs, fails to carry positional information in representations, and therefore has no intrinsic mechanism for representing token order. Existing Transformer architectures therefore explicitly inject positional information through mechanisms such as sinusoidal positional encoding~\citep{vaswani2017attention}, learned positional embeddings~\citep{devlin2019bert}, rotary positional encoding (RoPE)~\citep{su2021roformer}, and their long-context extensions such as YaRN~\citep{peng2023yarn}. While these approaches effectively break permutation equivariance and enable sequential modeling, positional information remains an externally introduced architectural component. Consequently, extending context lengths beyond those encountered during training typically requires additional adaptations to the positional encoding mechanism, rather than arising naturally from the architecture itself.

\begin{figure}[t]
  \centering
  
    \includegraphics[width=0.9\linewidth]{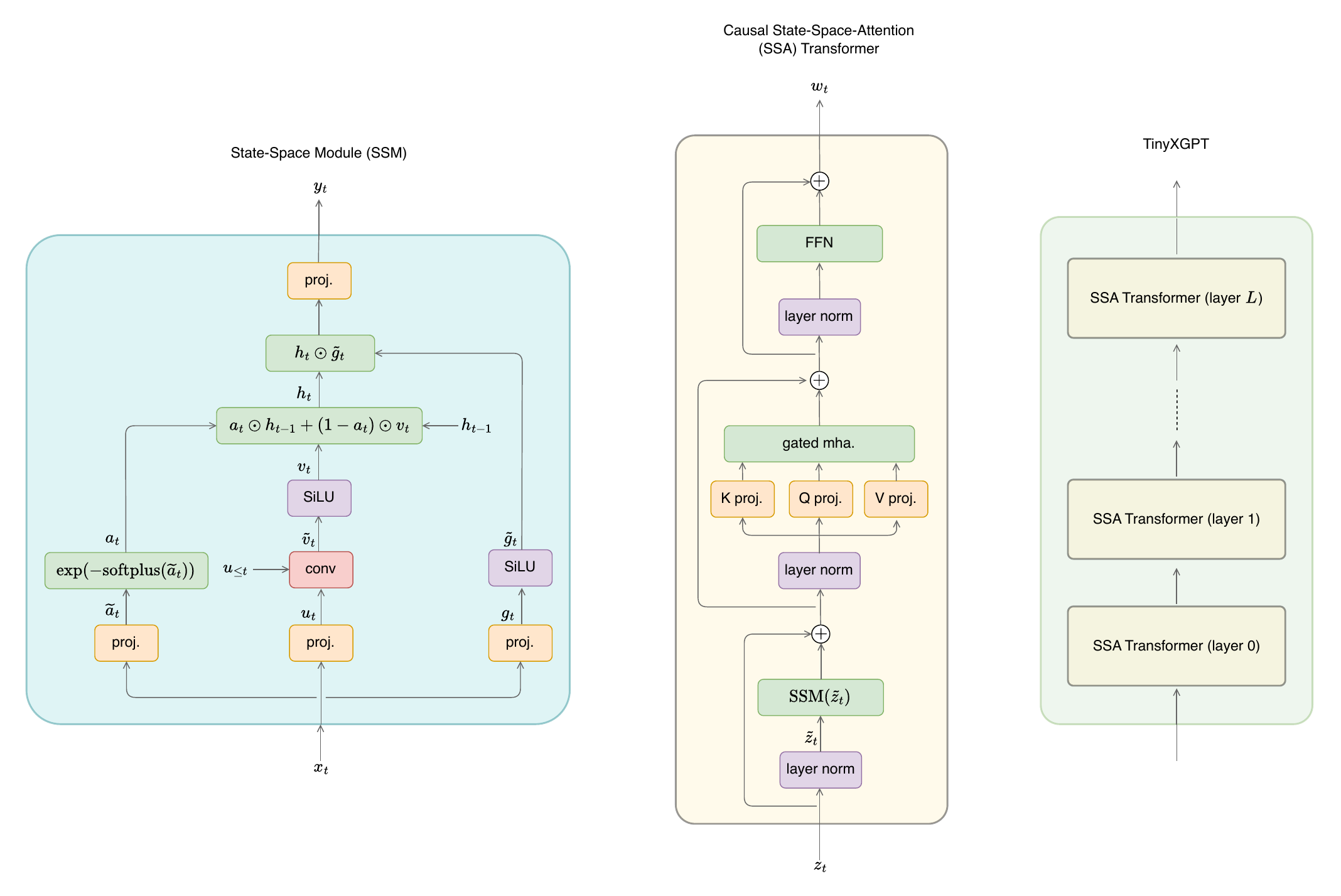}
  
    \caption{\textbf{Model Architecture.} \textsc{ZetaGPT} consists of a stack of State--Space--Attention (SSA) Transformer blocks (right). Each SSA Transformer block (middle) comprises three pre-normalized residual sub-layers: a causal state-space module (SSM), a gated multi-head self-attention layer, and a feed-forward network (FFN). The SSM precedes self-attention to produce position-aware representations through recurrent state evolution, allowing the subsequent attention layer to operate without explicit positional encodings. The left panel illustrates the internal architecture of the SSM, including the input-dependent state transition, recurrent state update, output gating, and output projection.}

  \label{fig:model}
\end{figure}

Although explicit positional encodings have been extended to longer contexts through interpolation and scaling techniques, such as position interpolation~\citep{chen2023extending}, LongRoPE~\citep{ding2024longrope}, and YaRN~\citep{peng2023yarn}, these approaches continue to rely on explicit positional encodings and adapt the encoding mechanism itself rather than the underlying architecture. This challenge has motivated a recent design philosophy shift toward positional-encoding-free architectures, in which positional information is represented implicitly through the computational dynamics of the architecture instead of being injected as an explicit prior. For example, Kimi Linear~\citep{kimi2025linear} eliminates explicit positional encodings by interleaving Kimi Delta Attention with full attention, allowing sequential information to emerge from the recurrent dynamics of the linear-attention module.

This design philosophy motivates this research, we incorporate causal state-space equations into Transformer blocks to produce position-aware representations through recurrent state evolution prior to self-attention. Consequently, positional information is provided implicitly by the recurrent dynamics, while self-attention retains its expressive capacity for modeling long-range token interactions. Formally, following selective state-space models~\citep{gu2023mamba}, given an input sequence $\{x_t\}_{t=1}^{T}$, the recurrent state evolution is defined as
\begin{align}
    h_t &= A(x_t)h_{t-1} + B(x_t)x_t,\\
    y_t &= C(x_t)h_t + D(x_t)x_t,
\end{align}
where $h_t$ denotes the hidden state at step $t$, $y_t$ is the output representation, and $A(\cdot)$, $B(\cdot)$, $C(\cdot)$, and $D(\cdot)$ denote the input-dependent state transition, input, output, and direct feed-through operators, respectively. Since the hidden state recursively summarizes the causal prefix $\{x_1,\ldots,x_t\}$, the resulting representations are inherently position-aware and are subsequently processed by the self-attention layer without requiring explicit positional encodings. Since the hidden state recursively summarizes the causal prefix $\{x_1,\ldots,x_t\}$, the resulting representations are inherently position-aware and are subsequently processed by the self-attention layer without requiring explicit positional encodings.

\begin{table}[t]
\centering
\small

\resizebox{0.86\linewidth}{!}{
\begin{tabular}{lcccccrrr}
\toprule
\textbf{Configuration} & \textbf{Layers} & \textbf{Heads} & \textbf{$d_{\mathrm{model}}$} &
\textbf{$d_h$} & \textbf{MLP} & \textbf{Embedding} & \textbf{Blocks} &
\textbf{Parameters} \\
\midrule
\textsc{ZetaGPT}-S (default) & $6$  & $8$  & $384$  & $48$ & $4\times$ & $19.3$M & $15.1$M  & $34.4$M \\
\textsc{ZetaGPT}-M           & $12$ & $12$ & $768$  & $64$ & $4\times$ & $38.6$M & $120.6$M & $159.2$M \\
\textsc{ZetaGPT}-L           & $24$ & $16$ & $1024$ & $64$ & $4\times$ & $51.5$M & $428.4$M & $479.9$M \\
\bottomrule
\end{tabular}
}

\caption{\textbf{\textsc{ZetaGPT} Configuration Scheme.} Every configuration scheme adopts the same State--Space--Attention architecture comprising a causal state-space module, gated multi-head self-attention, a feed-forward network, and no explicit positional encoding. Parameter counts assume the induced vocabulary size $V=50{,}259$, where the embedding parameters equal $Vd_{\mathrm{model}}$ and the block parameters equal $L(17d_{\mathrm{model}}^2+25d_{\mathrm{model}})$. Context length is a training configuration rather than an architectural constraint, since the model contains no explicit positional encoding.}

\label{tab:config_scheme}

\end{table}


We present \textsc{ZetaGPT}, a compact positional-encoding-free language model designed for research, rapid prototyping, algorithm verification, and educational applications. Beyond the proposed State--Space--Attention architecture, \textsc{ZetaGPT} provides a fully open-source, end-to-end training pipeline spanning dataset construction, tokenizer training, pretraining, supervised fine-tuning, reinforcement learning from human feedback (RLHF)~\citep{ouyang2022instructgpt}, and chain-of-thought (CoT) reasoning~\citep{wei2022chain} via pure reinforcement learning~\citep{guo2025deepseekr1}. To the best of our knowledge, \textsc{ZetaGPT} is the first open-source small language model without explicit positional encoding, establishing a compact and reproducible reference implementation for the development and empirical study of positional-encoding-free language models. The contributions of this work are summarized as follows:
\begin{enumerate}
    
    \item \textbf{Positional--Encoding--Free State--Space--Attention Language Model}. We present \textsc{ZetaGPT}, a compact positional-encoding-free State-Space--Attention language model in which causal state-space equations precede self-attention within every Transformer block. The recurrent state evolution produces position-aware representations prior to self-attention, enabling positional information to be represented implicitly through architectural dynamics rather than explicit positional encodings.

    \item \textbf{Reference Implementation for Positional-Encoding-Free Language Models}. We present \textsc{ZetaGPT} as a compact, fully open-source reference implementation for the development and empirical study of positional-encoding-free language models. To the best of our knowledge, it is the first open-source small language model without explicit positional encoding.

    \item \textbf{End-to-End LLM Pipeline}. We present, to the best of our knowledge, the first end-to-end pipeline for building modern small LLMs, encompassing every major stage from dataset construction, tokenizer training, pretraining, supervised fine-tuning, reinforcement learning from human feedback (RLHF), to chain-of-thought (CoT) reasoning via pure reinforcement learning. The pipeline provides a complete and reproducible recipe for research, rapid prototyping, algorithm verification, and education.
    
\end{enumerate}

\begin{table}[t]
\centering
\small
\setlength{\tabcolsep}{4.5pt}

\resizebox{0.96\linewidth}{!}{
\begingroup

\definecolor{baseline}{RGB}{245,243,255}
\definecolor{ours}{RGB}{236,254,255}

\setlength{\tabcolsep}{5pt}
\renewcommand{\arraystretch}{1.10}

\begin{tabular}{llrrlll}
\toprule
\textbf{Model} &
\textbf{Reference} &
\textbf{Params} &
\textbf{Vocab.} &
\textbf{Architecture} &
\textbf{Positional Encoding} &
\textbf{Pretrain Context} \\
\midrule

\multicolumn{7}{@{}l}{\textit{Positional-Encoding-Dependent Language Models}} \\

\rowcolor{baseline}
\textsc{TinyStories}-1M
& \citet{eldan2023tinystories}
& $3.7$M$^\dagger$
& $50{,}257$
& Transformer
& Learned
& 512 \\

\rowcolor{baseline}
Baby GPT (Character)
& \citet{karpathy2022nanogpt}
& $10.8$M$^\dagger$
& $65$
& Transformer
& Learned
& 256 \\

\rowcolor{baseline}
\textsc{TinyStories}-8M
& \citet{eldan2023tinystories}
& $19.7$M$^\dagger$
& $50{,}257$
& Transformer
& Learned
& 512 \\

\rowcolor{baseline}
\textsc{TinyStories}-33M
& \citet{eldan2023tinystories}
& $68.5$M$^\dagger$
& $50{,}257$
& Transformer
& Learned
& 512 \\

\rowcolor{baseline}
Pythia-70M
& \citet{biderman2023pythia}
& $70.4$M
& $50{,}304$
& Transformer
& RoPE
& 2048 \\

\rowcolor{baseline}
GPT-2 Small
& \citet{radford2019gpt2}
& $124$M
& $50{,}257$
& Transformer
& Learned
& 1024 \\

\rowcolor{baseline}
SmolLM2-135M
& \citet{allal2025smollm2}
& $134.5$M
& $49{,}152$
& Transformer
& RoPE
& 8192 \\

\rowcolor{baseline}
Gemma 3 270M
& \citet{google2025gemma3270m}
& $268.1$M
& $262{,}144$
& Transformer
& RoPE
& 32768 \\

\rowcolor{baseline}
nanochat d20
& \citet{karpathy2025nanochat}
& $\sim560$M
& $65{,}536$
& Transformer
& RoPE
& 1024 \\

\rowcolor{baseline}
Qwen3-0.6B
& \citet{qwen2025qwen3}
& $\sim0.6$B
& $151{,}936$
& Transformer
& RoPE
& 32768 \\

\rowcolor{baseline}
\textsc{TinyLlama}
& \citet{zhang2024tinyllama}
& $\sim1.1$B
& $32{,}000$
& Transformer
& RoPE
& 2048 \\

\midrule
\multicolumn{7}{@{}l}{\textit{Ours: Positional-Encoding-Free Language Models}} \\

\rowcolor{ours}
\textsc{ZetaGPT}-S (default)
& ---
& $34.4$M
& $50{,}259$
& \textbf{State--Space--Attention}
& \textbf{None}
& \textbf{256} \\

\rowcolor{ours}
\textsc{ZetaGPT}-M
& ---
& $159.2$M
& $50{,}259$
& \textbf{State--Space--Attention}
& \textbf{None}
& \textbf{512} \\

\rowcolor{ours}
\textsc{ZetaGPT}-L
& ---
& $479.9$M
& $50{,}259$
& \textbf{State--Space--Attention}
& \textbf{None}
& \textbf{1024} \\

\bottomrule
\end{tabular}

\endgroup
}

\caption{\textbf{\textsc{ZetaGPT}'s Niche Among Compact Language Models.} Parameter counts are reported from the corresponding papers, repositories, or official model cards; $\dagger$ denotes values derived from released configurations when an explicit parameter count was not reported. \emph{Architecture} indicates the underlying model architecture, while \emph{Positional Encoding} specifies the mechanism used to represent token order. \textsc{ZetaGPT} occupies a distinct position as the only compact State--Space--Attention language model without explicit positional encoding, providing a reference implementation for positional-encoding-free language modeling.}

\label{tab:landscape}
\end{table}

\section{Landscape of Small Language Models}
\label{sec:landscape}

Table~\ref{tab:landscape} positions \textsc{ZetaGPT} among representative compact and sub-billion language models. Existing small language models are almost exclusively based on the Transformer architecture and represent token order through explicit positional encoding, either using learned positional embeddings, as in GPT-2~\citep{radford2019gpt2} and the \textsc{TinyStories} family~\citep{eldan2023tinystories}, or rotary positional encoding (RoPE), as adopted by more recent models including Pythia~\citep{biderman2023pythia}, SmolLM2~\citep{allal2025smollm2}, Gemma~3~\citep{google2025gemma3270m,gemma2025gemma3}, Qwen3~\citep{qwen2025qwen3}, and \textsc{TinyLlama}~\citep{zhang2024tinyllama}. Although these models differ substantially in scale and training objectives, they all rely on explicit positional representations as an architectural component.

In contrast, \textsc{ZetaGPT} adopts a hybrid State--Space--Attention Transformer architecture without explicit positional encoding. Instead of injecting positional information into token representations, sequential information is represented implicitly through recurrent state evolution before self-attention. To the best of our knowledge, \textsc{ZetaGPT} is the first open-source small language model that combines a positional-encoding-free State--Space--Attention architecture with a complete end-to-end training pipeline. This positions \textsc{ZetaGPT} as a compact reference implementation for the development, evaluation, and empirical study of positional-encoding-free language models.

\section{Model}
\label{sec:model}

As shown in Figure~\ref{fig:model}, \textsc{ZetaGPT} adopts a hybrid State--Space--Attention architecture in which a causal state-space module precedes self-attention within every Transformer block. The state-space module first encodes sequential information through recurrent state evolution, producing position-aware representations that are subsequently processed by self-attention without requiring explicit positional encodings. Each block consists of three pre-normalized residual sub-layers,
\begin{align}
  x &\leftarrow x + \mathrm{SSM}\big(\mathrm{LN}(x)\big), \qquad
  x \leftarrow x + \mathrm{Attn}\big(\mathrm{LN}(x)\big), \qquad
  x \leftarrow x + \mathrm{MLP}\big(\mathrm{LN}(x)\big),
  \label{eq:block}
\end{align}
where the state-space module models sequential dynamics, the attention layer captures global token interactions, and the feed-forward network performs nonlinear feature transformation. 

Table~\ref{tab:config_scheme} summarizes the three model configurations. \textsc{ZetaGPT}-S is the default configuration used throughout this work, comprising 6 State--Space--Attention Transformer blocks with 8 attention heads, a model dimension of 384, a head dimension of 48, a $4\times$ feed-forward expansion, and 34.4M parameters. \textsc{ZetaGPT}-M scales the architecture to 12 layers, 12 attention heads, a model dimension of 768, and 159.2M parameters. \textsc{ZetaGPT}-L further scales the architecture to 24 layers, 16 attention heads, a model dimension of 1024, and 479.9M parameters. All three configurations share the same architectural design and differ only in model depth and width.

\begin{figure}[t]
  \centering
  \includegraphics[width=\linewidth]{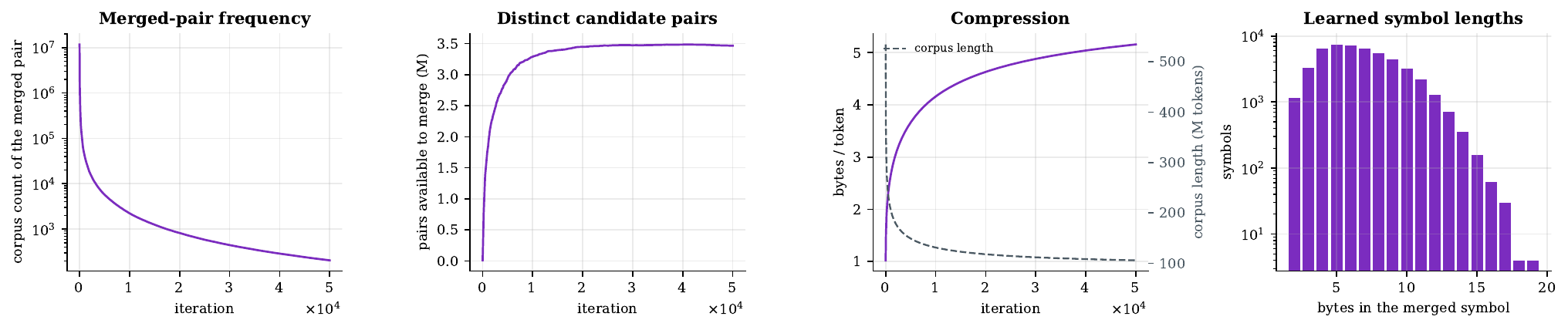}
  
  \caption{\textbf{Dynamics of BPE Tokenizer.} This shows the dynamics of BPE tokenizer over $50{,}000$ merge iterations on the full
  $546$\,MB Wikitext-103 corpus, one merge per iteration. \emph{Left:} the corpus frequency of the pair
  being merged falls from $1.2\times10^{7}$ to $205$, almost five orders of magnitude (note
  the logarithmic axis), so late merges are learned from very little evidence.
  \emph{Centre-left:} the number of distinct adjacent pairs available to merge \emph{rises}
  by a factor of $365$, from $9{,}497$ to $3.47$M, reaching $90\%$ of its final value by
  iteration ${\sim}6{,}700$: merging creates more adjacencies than it consumes, which is why
  the candidate set has to be maintained incrementally rather than recounted.
  \emph{Centre-right:} compression rises monotonically from $1.02$ to $5.16$ bytes per token
  and the corpus falls from $535$M to $106$M tokens, but with sharply diminishing returns ---
  the first $1{,}000$ iterations buy $+1.63$ bytes/token, the next $9{,}000$ buy $+1.51$, and
  the final $40{,}000$ only $+1.00$. \emph{Right:} the byte length of the symbol each merge
  creates, over the whole run; the mean rises from $2.35$ bytes in the first hundred
  iterations to $7.19$ in the last thirty thousand, with the longest at $19$ bytes.}
  
  \label{fig:bpe}
\end{figure}

\subsection{Tokenizer}
\label{sec:tokenizer}

\textsc{ZetaGPT} employs a byte-level byte-pair encoding (BPE) tokenizer~\citep{sennrich2016bpe}, following the tokenization paradigm adopted by GPT-2~\citep{radford2019gpt2}. Starting from the 256-byte vocabulary, the tokenizer iteratively merges the most frequent adjacent symbol pairs learned from the training corpus, producing a subword vocabulary that balances compression efficiency and lexical coverage. Let $\mathcal{M}$ denote the set of learned merge rules. The resulting vocabulary size is
\begin{align}
    V = 259 + |\mathcal{M}|,
\end{align}
where the additional three symbols correspond to the special tokens $\{\langle\mathrm{pad}\rangle,\langle\mathrm{eos}\rangle,\langle\mathrm{unk}\rangle\}$. Since the base alphabet consists of raw bytes rather than words or Unicode characters, every input string admits a valid encoding without out-of-vocabulary failures.

A single tokenizer is trained over the pretraining corpus. Unless otherwise specified, the merge budget is fixed to $|\mathcal{M}|=50{,}000$, resulting in a vocabulary size of $V=50{,}259$, which is used by all model configurations reported in this work. Figure~\ref{fig:bpe} illustrates the tokenizer learning dynamics throughout the byte-pair merging process, including the evolution of merge frequency, vocabulary compression, and corpus tokenization efficiency. 

\begin{figure}[t]
  \centering
  \includegraphics[width=\linewidth]{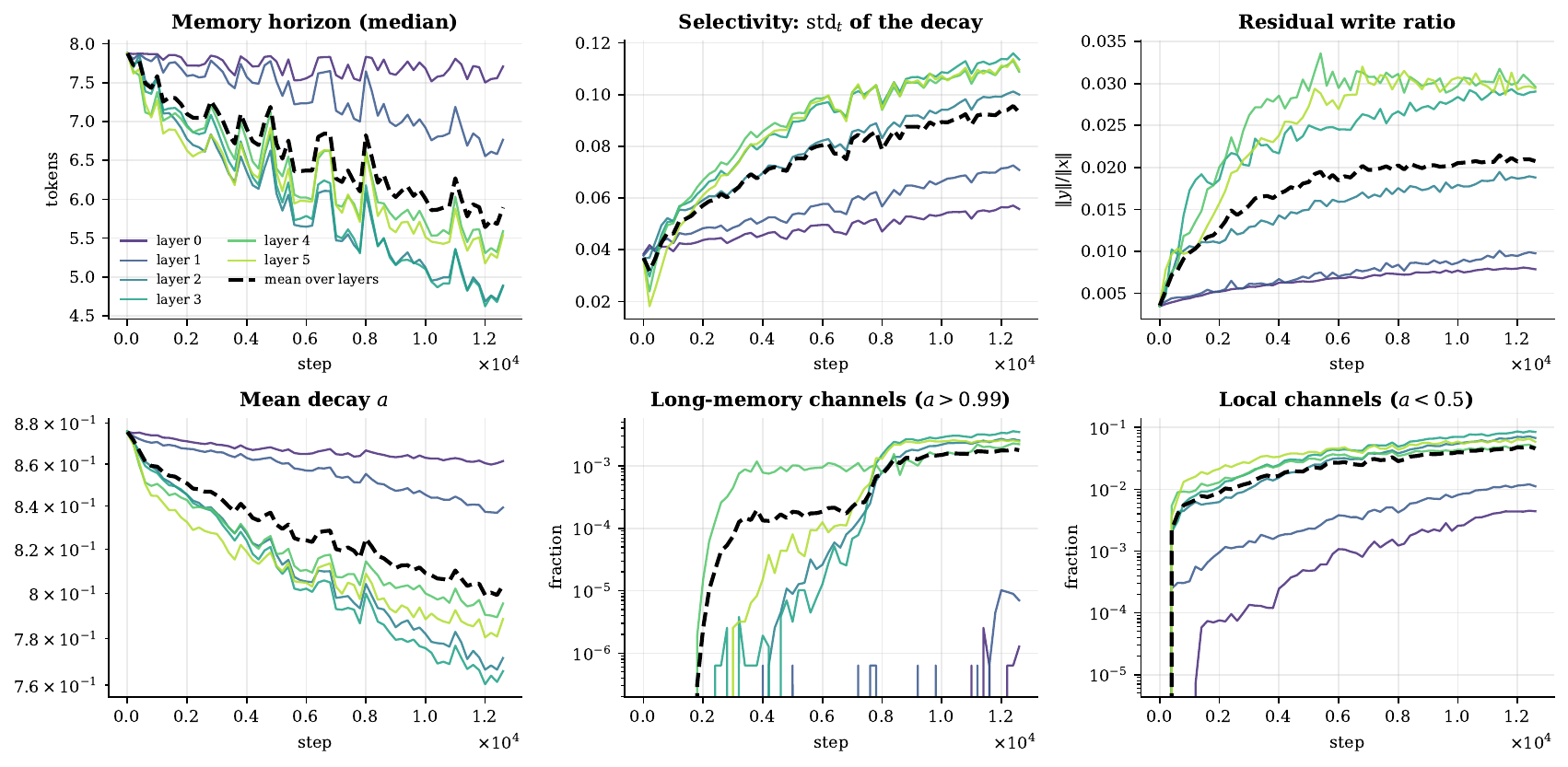}
  
    \caption{\textbf{Per-Layer Dynamics of the State-Space Modules.} Training dynamics of the state-space modules, profiled every 200 optimization steps for each Transformer block. Curves are shown from shallow (light) to deep (dark), with the dashed curve indicating the mean across all blocks. The learned dynamics progressively become more selective: the memory horizon shortens, the input-dependent decay becomes increasingly adaptive, and the contribution of the state-space module to the residual stream grows throughout training. The bottom row further shows the evolution of the decay distribution, illustrating the emergence of both long-memory and short-memory channels. \textbf{Together, these dynamics demonstrate that the recurrent state evolution learns multiple memory timescales, providing the mechanism through which sequential information is represented implicitly before self-attention.}}
  
  \label{fig:ssm}
\end{figure}


\subsection{State-Space Module}
\label{sec:ssm}

Discrete state-space equations define the input--output dynamics of sequential data through latent hidden variables whose evolution is governed recursively across discrete sequence steps. This discrete dynamical evolution inherently carries sequential order: the hidden state at step $t$ depends on the state propagated from preceding steps and therefore represents the causal history associated with its position in the sequence. Accordingly, state-space dynamics provide an architectural mechanism for encoding positional information without introducing an explicit positional representation. Following state-space and selective state-space formulations~\citep{gu2022s4,gu2023mamba}, the dynamics are written as
\begin{align}
    h_t &= A(x_t)h_{t-1} + B(x_t)x_t, \\
    y_t &= C(x_t)h_t + D(x_t)x_t,
    \label{eq:ssm_general}
\end{align}
where $x_t$ is the input at step $t$, $h_t$ is the latent state, $y_t$ is the output representation, and $A(\cdot)$, $B(\cdot)$, $C(\cdot)$, and $D(\cdot)$ denote the state-transition, input, output, and direct feed-through operators, respectively. In the selective formulation, these operators may depend on the current input, allowing the state dynamics to adapt to the sequence content.

\textsc{ZetaGPT} instantiates this formulation as a selective diagonal state-space module. The state-transition operator is defined as
\begin{align}
    A(x_t) &= \operatorname{diag}(a_t), \qquad
    a_t = \exp\!\left(-\operatorname{softplus}(W_a x_t + b_a)\right),
    \label{eq:ssm_decay}
\end{align}
where $a_t\in(0,1)^d$ is an input-dependent per-channel decay. The input path first constructs a value representation through a learned projection, depthwise causal convolution, and nonlinearity,
\begin{align}
    u_t &= W_v x_t + b_v, \\
    v_t &= \operatorname{SiLU}\!\left(\operatorname{Conv}_{\mathrm{causal}}(u)_t\right),
\end{align}
while a parallel projection produces the output gate
\begin{align}
    g_t &= W_g x_t + b_g.
\end{align}
The operators in~\eqref{eq:ssm_general} are then instantiated as
\begin{align}
    A(x_t)h_{t-1} &\equiv a_t\odot h_{t-1}, \\
    B(x_t)x_t &\equiv (1-a_t)\odot v_t, \\
    C(x_t)h_t &\equiv W_o\!\left(h_t\odot\operatorname{SiLU}(g_t)\right)+b_o, \\
    D(x_t)x_t &\equiv 0,
    \label{eq:ssm_mapping}
\end{align}
where $\odot$ denotes element-wise multiplication. Substitution into the general state-space equations yields the instantiated recurrence
\begin{align}
    h_t &= a_t\odot h_{t-1} + (1-a_t)\odot v_t, \\
    y_t &= W_o\!\left(h_t\odot\operatorname{SiLU}(g_t)\right)+b_o.
    \label{eq:ssm_inst}
\end{align}

Because $h_t$ is obtained recursively from $h_{t-1}$, the representation at each step is conditioned on the ordered causal prefix rather than solely on the current token. The state-space module therefore transforms position-agnostic token representations into position-aware representations before self-attention, providing the sequential information required by the subsequent attention computation without explicit positional encoding.

\paragraph{Observed State-Space Dynamics.}
The state-space equations are introduced in \textsc{ZetaGPT} to represent positional information implicitly through recurrent dynamics without explicit positional encodings. Figure~\ref{fig:ssm} reports the evolution of the state-space dynamics during the first $12{,}800$ pretraining steps, corresponding to $19.7\%$ of the total optimization budget, over which the training loss decreases from $10.90$ to $6.05$ nats per token. The reported quantities are proxy measurements of the learned recurrent dynamics and characterize how positional information is represented across layers throughout optimization. Collectively, the dynamics progressively develop multiple memory timescales, allowing different channels to specialize to distinct temporal ranges and thereby enabling the state-space module to encode sequential information before self-attention. Four consistent observations support this interpretation.

\begin{itemize}

    \item \textbf{Emergence of Multi-Scale Memory Horizons.}
    The median memory horizon, defined as $\tau=-(\ln a_t)^{-1}$, where $a_t \in (0,1)$ is the input-dependent state decay coefficient of the state-space module, decreases from approximately $7.9$ to $5.7$ tokens before stabilizing near the end of the observation window. More importantly, memory horizons progressively differentiate across layers rather than converging to a common value. The shallowest layers consistently retain the longest memory horizons, the intermediate layers develop the shortest horizons, and the deepest layers converge to intermediate horizons. This layer-wise specialization indicates that different state-space modules become sensitive to sequential dependencies over distinct temporal ranges, thereby forming a hierarchy of memory horizons across the network.

    \item \textbf{Growth of Input Selectivity.}
    Selectivity, measured as the standard deviation of the input-dependent decay across token positions, increases from $0.037$ to $0.094$, indicating that the state-space dynamics become progressively more input-dependent throughout optimization. Rather than converging toward a fixed exponential filter, different tokens increasingly induce distinct decay coefficients, allowing the memory horizon to adapt to the local input context. The strongest selectivity is consistently observed in the intermediate layers, coinciding with the greatest reduction in memory horizon.

    \item \textbf{Increasing Contribution to the Residual Stream.}
    The residual write ratio, defined as $\|f(x)\|_F/\|x\|_F$, where $x$ is the normalized input presented to the state-space module and $f(x)$ denotes the state-space output prior to residual addition, both norms being computed over the full activation tensor of a training batch, increases by approximately a factor of six, from $0.004$ to $0.021$. This trend indicates that the network progressively relies more heavily on the state-space module during optimization. The increase is also layer-dependent, with deeper blocks exhibiting substantially larger write ratios than shallower ones. Although the state-space branch contributes only a modest fraction of the residual activation, its contribution continues to increase throughout the observation window without clear evidence of convergence, suggesting that the learned state-space representations become increasingly important to the overall model computation.

    \item \textbf{Emergence of Multiple Memory Timescales.}
    The distribution of decay coefficients evolves from a narrow initialization into a heterogeneous population containing both short-memory and long-memory channels. Long-memory channels ($a>0.99$) gradually emerge after approximately $2\times10^3$ optimization steps, whereas short-memory channels ($a<0.5$) appear much earlier and eventually constitute several percent of the channel population. This heterogeneous distribution is learned rather than prescribed by the architecture, demonstrating that the state-space modules automatically organize into multiple memory timescales. Together with the layer-wise specialization of memory horizons, increased input selectivity, and growing contribution to the residual stream, these observations indicate that the state-space dynamics progressively learn a hierarchy of temporal representations, providing the architectural mechanism through which positional information is represented implicitly before self-attention.

\end{itemize}

\subsection{Gated Multi-Head Attention}
\label{sec:gated}


Attention sink, in which disproportionate attention is assigned to a small number of semantically uninformative tokens, is a widely observed phenomenon in autoregressive Transformer language models~\citep{gu2025attentionsink}. Recent studies further suggest that attention sink and related attention-head behaviors are closely associated with hallucination and that improving attention allocation can substantially mitigate hallucinated generations~\citep{yuan2024faith,chuang2024lookback}. Gating the attention output has been shown to improve attention selectivity while mitigating attention-sink and activation-collapse phenomena in pre-layer-normalized Transformers~\citep{qiu2025gatedattention}. Accordingly, \textsc{ZetaGPT} adopts the gated multi-head attention mechanism proposed by \citeauthor{qiu2025gatedattention}~\citep{qiu2025gatedattention} while retaining the standard causal Transformer attention formulation~\citep{vaswani2017attention}.

Given an input sequence $x\in\mathbb{R}^{T\times d_{\mathrm{model}}}$, the query, key, and value representations are computed as
\begin{align}
    Q=xW_Q,\qquad
    K=xW_K,\qquad
    V=xW_V,
\end{align}
where $W_Q$, $W_K$, and $W_V$ denote the learnable query, key, and value projection matrices, respectively. The projected representations are partitioned into $H$ attention heads with head dimension $d_h=d_{\mathrm{model}}/H$. For each attention head,
\begin{align}
    A^{(h)}
    &=
    \operatorname{softmax}\!\left(
        \frac{Q^{(h)}K^{(h)\top}}{\sqrt{d_h}}
        +M
    \right),\\
    O^{(h)}
    &=
    A^{(h)}V^{(h)},
\end{align}
where $M$ denotes the causal attention mask. The outputs from all attention heads are concatenated to form
\begin{align}
    O=[\,O^{(1)},O^{(2)},\ldots,O^{(H)}\,].
\end{align}

Instead of directly projecting the concatenated attention output, the gated attention mechanism introduces an input-dependent gate,
\begin{align}
    \operatorname{Attn}(x)
    =
    \left(
        O\odot
        \sigma(xW_G)
    \right)
    W_O,
    \label{eq:gated_attn}
\end{align}
where $W_G$ and $W_O$ denote the gating and output projection matrices, respectively, and $\sigma(\cdot)$ is the sigmoid activation. The gating mechanism adaptively modulates the contribution of each attention channel according to the current token representation, introducing a nonlinear interaction between the attention output and the residual stream. As demonstrated by \citeauthor{qiu2025gatedattention}~\citep{qiu2025gatedattention}, such input-dependent gating improves attention selectivity while mitigating attention-sink and activation-collapse phenomena in pre-layer-normalized Transformers.

\section{Pipeline \& Data}
\label{sec:data}

\begin{figure}[t]
  \centering
  
    \includegraphics[width=0.86\linewidth]{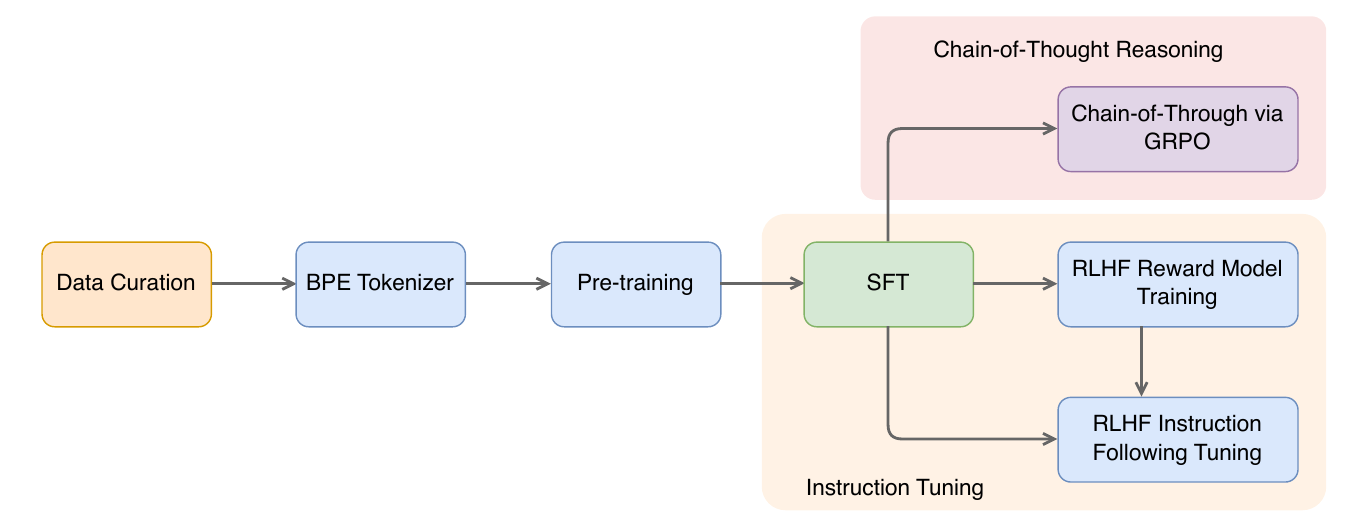}
  
    \caption{\textbf{End-to-End Training Pipeline.} The pipeline begins with data curation and tokenizer training, followed by language-model pretraining on the pretraining corpus. The pretrained model is subsequently aligned through supervised fine-tuning (SFT) on instruction-following data. A reward model is then initialized from the SFT model by replacing the language modeling head with a scalar reward head and trained on preference data for reinforcement learning from human feedback (RLHF). Finally, chain-of-thought (CoT) reasoning is further developed through reinforcement learning using Group Relative Policy Optimization (GRPO), reproducing the emergent ``\emph{aha moment}'' observed in recent reasoning language models~\citep{guo2025deepseekr1}.}
  
  \label{fig:pipeline}
  
\end{figure}

\paragraph{End-to-End Training Pipeline.} Figure~\ref{fig:pipeline} illustrates the complete end-to-end training pipeline of \textsc{ZetaGPT}. Starting from raw corpora, the pipeline sequentially performs tokenizer training, language-model pretraining, supervised fine-tuning (SFT), reward model training, reinforcement learning from human feedback (RLHF), and chain-of-thought (CoT) reasoning through reinforcement learning. Each stage is independently executable while remaining fully compatible with the subsequent stages, providing a unified and reproducible workflow for constructing modern small language models.

\paragraph{Pretraining Data.}
Language-model pretraining is performed on the WikiText-103 corpus~\citep{merity2017pointer}, a large-scale English language modeling benchmark comprising long-form Wikipedia articles. The corpus provides a clean and coherent pretraining source for learning general language modeling capabilities.

\paragraph{Instruction-Tuning Data.}
Supervised instruction tuning is performed using the Alpaca-GPT4 instruction-following dataset~\citep{peng2023instruction}, which consists of GPT-4-generated instruction--response pairs spanning a broad range of general-purpose tasks. The same instruction corpus is subsequently reused for reward model training and reinforcement learning from human feedback (RLHF), providing a unified supervision source throughout the alignment pipeline.

\paragraph{Chain-of-Thought Reasoning Data.}
Chain-of-thought (CoT) reasoning is developed through reinforcement learning using the GSM8K mathematical reasoning benchmark~\citep{cobbe2021gsm8k}. Following recent reasoning language models, optimization is performed using Group Relative Policy Optimization (GRPO)~\citep{shao2024deepseekmath}, enabling reasoning capability to emerge directly through reinforcement learning without supervised chain-of-thought demonstrations.

\section{Training Protocol}
\label{sec:training_protocol}

All training stages are optimized using AdamW with decoupled weight decay~\citep{loshchilov2019decoupled}, momentum coefficients $\beta=(0.9,0.999)$, $\varepsilon=10^{-8}$, gradient clipping with a maximum global norm of $1.0$, and a cosine annealing learning-rate schedule with the minimum learning rate fixed to one tenth of the peak learning rate. Language-model pretraining is performed with a learning rate of $2\times10^{-5}$ for $64{,}840$ optimization steps, followed by supervised fine-tuning (SFT) using a learning rate of $1\times10^{-6}$ for $2{,}342$ steps. Reward model training adopts a learning rate of $1\times10^{-5}$ for $2{,}500$ steps, while reinforcement learning from human feedback (RLHF) and chain-of-thought (CoT) reasoning through Group Relative Policy Optimization (GRPO) are optimized using a learning rate of $1\times10^{-6}$ for $3{,}251$ and $1{,}400$ optimization steps, respectively. Model checkpoints, training statistics, and state-space diagnostics are recorded every $200$ optimization steps. Unless otherwise specified, all stages use a batch size of $16$ sequences (or preference pairs), and the context length follows the pretraining configuration of the corresponding model.

\section{Limitations}

The limitations of \textsc{ZetaGPT} primarily arise from the computational resources available for this work rather than from the proposed architecture itself. First, the pretraining corpus is intentionally modest in scale and therefore cannot provide the linguistic diversity, world knowledge, and task coverage of contemporary large language models trained on trillions of tokens. Second, the model is deliberately compact, with the default configuration containing only $34.4$M parameters. Although this scale is well suited to research, rapid prototyping, algorithm verification, and education, conclusions drawn from experiments at this scale may not directly transfer to substantially larger language models whose optimization dynamics and emergent capabilities differ qualitatively. Third, computational constraints limit pretraining to relatively short context lengths (256, 512, and 1024 tokens for the three model configurations), preventing a comprehensive empirical evaluation of long-context modeling and context-length extrapolation. Evaluating the proposed positional-encoding-free architecture under substantially longer training contexts remains an important direction for future work.

\section{Conclusion}

This work presented \textsc{ZetaGPT}, a compact positional-encoding-free State--Space--Attention language model in which positional information is represented implicitly through state-space dynamics rather than explicit positional encodings. By placing a causal state-space module before self-attention within every Transformer block, \textsc{ZetaGPT} enables position-aware representations to emerge through learned state-space dynamics while preserving the expressive modeling capability of self-attention. Beyond the proposed architecture, \textsc{ZetaGPT} provides a complete end-to-end language model development pipeline spanning dataset construction, tokenizer training, pretraining, supervised fine-tuning, reward model training, reinforcement learning from human feedback (RLHF), and chain-of-thought (CoT) reasoning through reinforcement learning. To the best of our knowledge, \textsc{ZetaGPT} is the first open-source small language model without explicit positional encoding, establishing a compact and reproducible reference implementation for the development, evaluation, and empirical study of positional-encoding-free language models.

\bibliographystyle{plainnat}
\bibliography{references}

\end{document}